# IFRA: A Machine Learning-Based Instrumented Fall Risk Assessment Scale Derived from An Instrumented Timed Up and Go Test in Stroke Patients

**Simone Macciò [1], Alessandro Carfì [2], Alessio Capitanelli [1,*], Peppino Tropea [3], Massimo Corbo [4], Fulvio Mastrogiovanni [2] and Michela Picardi [5]**

[1] Teseo Srl, P.zza Nicolò Montano 2A/1, 16151 Genoa, Italy;

[2] Department of Informatics, Bioengineering, Robotics and Systems Engineering, University of Genoa, Viale Causa 13, 16145 Genoa, Italy;

[3] CantoniLab, Via Giovanni Cantoni 7, 20144 Milan, Italy;

[4] Need Institute, Foundation for the Cure and Rehabilitation of Neurological Diseases, 20121 Milan, Italy;

[5] Department of Neurorehabilitation Sciences, Casa Di Cura Igea, Via Dezza 48, 20144 Milan, Italy;

* Correspondence: alessio.capitanelli@dibris.unige.it

**Abstract**

**Background/Objectives:** Falls represent a major health concern for stroke survivors, necessitating effective risk assessment tools. This study proposes the Instrumented Fall Risk Assessment (IFRA) scale, a novel screening tool derived from Instrumented Timed Up and Go (ITUG) test data, designed to capture mobility measures often missed by traditional scales. **Methods**: We employed a two-step machine learning approach to develop the IFRA scale: first, identifying predictive mobility features from ITUG data and, second, creating a stratification strategy to classify patients into low-, medium-, or high-fall-risk categories. This study included 142 participants, who were divided into training (including synthetic cases), validation, and testing sets (comprising 22 non-fallers and 10 fallers). IFRA's performance was compared against traditional clinical scales (e.g., standard TUG and Mini-BESTest) using Fisher's Exact test. **Results**: Machine learning analysis identified specific features as key predictors, namely vertical and medio-lateral acceleration, and angular velocity during walking and sit-to-walk transitions. IFRA demonstrated a statistically significant association with fall status (Fisher's Exact test $p$ = 0.004) and was the only scale to assign more than half of the actual fallers to the high-risk category, outperforming the comparative clinical scales in this dataset. **Conclusions**: This proof-of-concept study demonstrates IFRA's potential as an automated, complementary approach for fall risk stratification in post-stroke patients. While IFRA shows promising discriminative capability, particularly for identifying high-risk individuals, these preliminary findings require validation in larger cohorts before clinical implementation.

# 1. Introduction

Falls are the second leading cause of unintentional injury deaths worldwide, after road traffic injuries [1]. Individuals who have experienced a stroke are seven times more likely to experience a fall compared to healthy individuals [2–5]. Within the first year following a stroke, the likelihood of a fall is estimated at 50% [6], with 5% of these falls leading to serious injuries [7,8]. Falls also cause anxiety, fear of falling, and depression, thereby reducing independence and further triggering a vicious cycle of inactivity, which increases disability and the risk of falling itself [9,10]. Fall risk assessments and interventions to reduce the risk of falls are therefore needed to reduce this burden.

Effective interventions for mitigating the risk of falling include rehabilitation (which involves therapeutic exercise), home adaptations, and education on fall risk factors. Assessing the number and severity of a patient's fall risk factors is essential, since many of these factors could be, at least in principle, acted upon (for example, poor balance or home hazards for falls). The list of fall risk factors in stroke is long, with over 100 risk factors reported in a recent systematic review [11]. Fall risk charts for stroke patients include heterogeneous variables [12], such as advanced age, previous strokes or falls, hemiparesis severity (that is, the degree of mobility impairment), the presence of cognitive impairment or urinary incontinence, the use of sedatives and other psychotropic drugs, as well as the presence of depression [13].

Because of the multiple factors contributing to falls [14–17], fall risk identification remains a challenge. Identifying an effective screening tool, as well as a concise and easy-to-use measure, is crucial, as there are various evidence-based interventions that can support individuals who are at risk and could benefit from preventive strategies [18].

Among the most widely adopted screening tools is the Timed Up and Go (TUG) assessment [19–21]. The time needed to complete the test is commonly measured with a stopwatch, and usually the TUG test duration (total TUG duration, TTD) represents the only measurement extracted from this test. The TUG test is chosen because it is simple, quick, and evaluates several key risk factors, including gait and balance, in a single assessment. In fact, the TUG is considered a functional mobility measure, since it incorporates dynamic balance tasks, including sit-to-stand and stand-to-sit transitions, turning, and straight-line walking [20–24]. Although the TUG relies solely on total time as its outcome measure, this metric has demonstrated some ability to distinguish fallers from non-fallers, albeit inconsistently [20–24].

In order to enhance the TUG's ability to assess fall risk, researchers developed the instrumented Timed Up and Go (ITUG), which integrates wearable inertial measurement units (IMUs) [25]. Indeed, in motor disorder assessment, IMUs can broaden traditional clinical tests into a source of novel mobility measures. Several classical mobility tests have been extended through the use of IMUs; among these is the ITUG test [25]. The ITUG provides detailed quantitative information on each TUG sub-task (that is, sit-to-walk, turn, turn-to-sit), such as durations, accelerations, and angular velocities. The ITUG test can identify distinct gait patterns and balance impairments [26], is validated against standard kinematic measures [27], and can distinguish between individuals who have experienced falls and those who have not [28]. In addition to research applications, commercial ITUG systems are now available, offering user-friendly software that enables quick administration and automatic reporting of test outcomes [29]. In patients with neurological diseases, measures derived from the ITUG have shown sound psychometric properties, such as appropriate criterion and construct validity and responsiveness to rehabilitation [30–32]. A current challenge associated with this additional information is that, as more outputs become available to describe assessment performance, there is a lack of evidence regarding which parameters are most valuable for fall risk screening.

In our previous work [30–32], in which a diverse cohort of patients with neurological diseases (including stroke) was longitudinally assessed, we showed that single measures from the ITUG test, obtained using a single IMU secured to the patient's trunk, have some diagnostic ability to identify patients at risk of falling. However, in strict diagnostic terms, the accuracy of these measures can still be improved.

Based on this background, the aim of the current study is to create a new fall risk assessment scale, which we refer to as the "Instrumented Fall Risk Assessment" (IFRA). This new measure leverages features extracted from the ITUG and allows clinicians to categorize patients into high-, medium-, and low-risk fall groups. Finally, we compare the categorization results obtained using IFRA against the metrics currently used by clinicians to assess patients' fall risk in clinical practice.

## 2. Data

The current work relies on data already presented by Caronni et al., 2023 [33], which was collected at the Neurorehabilitation Department of Casa di Cura Igea (CCI; Milan, Italy) between October 2018 and January 2020. All participants gave their written informed consent, and the study was approved by the local ethical committee (Comitato Etico Milano Area 2; 568_2018bis).

### *2.1. Participants Selection*

Building on the dataset already presented in [33], our study includes only patients older than 18 years who were affected by an ischemic or hemorrhagic cortical or subcortical stroke with gait (that is, hemiparetic gait) and balance impairments. These patients are able to walk and move from sitting to standing without assistance. We exclude participants with any acute medical conditions or other conditions that could impair mobility on their own, such as severe heart failure.

### *2.2. Participants Assessment*

Similarly to prior research on balance measure validity [30,32], we collected clinical and instrumental movement measures alongside relevant clinical data. Each participant underwent assessments by a physiotherapist or an occupational therapist, who also conducted the instrumented mobility assessment. Data collection occurred at the time of the patient's discharge from the rehabilitation unit. During the assessment, we prioritized testing without gait aids. However, gait aids were allowed at the clinician's discretion if the fall risk during testing was too high. Similarly, ankle foot orthoses were permitted if needed.

Below, we briefly summarize the assessment battery that includes the clinical (that is, usually adopted in the clinical practice) and the instrumented assessment (that is, regarding the Instrumented Timed Up and Go test) administered in the study to the patient's sample.

#### *2.2.1. Clinical Assessment*

*Timed Up and Go (TUG) test*: The TUG test [21] is described as the total time in seconds (total time duration, TTD), measured by a stopwatch, that a subject takes to rise from a chair, walk 3 m, turn 180 degrees, return to the chair, turn again 180 degrees, and sit down. To ensure reliable test results, the TUG test was repeated five times for each participant, and the average time was considered. Participants were instructed to maintain a comfortable and safe walking speed during all repetitions.

*10 m walking test (10 MWT):* 10 MWT [34] is the comfortable speed that a subject takes to walk a straight 10 m path at their preferred pace. Their walking speed is calculated based on the time to walk the middle 6 m, excluding starting (that is, time of acceleration) and stopping (that is, time to deceleration). As for the TUG test, we considered the average value over five repetitions of the test.

*Mini-BESTest (MB):* MB [35–37] is a 14-item assessment tool that evaluates various aspects of balance impairment, including static standing balance and dynamic balance during ambulation.

*Functional Independence Measure (FIM)*: The FIM [38] is a standardized tool to assess the level of disability in performing activities of daily living (ADLs). FIM has two main domains, namely motor and cognitive.

*Performance Oriented Mobility assessment (POMA-B)*: POMA-B [38] is a widely used clinical scale for balance impairment *in the elderly population*.

*Conley Scale*: The Conley Scale [39] is a widely used fall risk clinical tool developed for use by nurses in hospital settings. It is a very quick scale to administer (approximately 5 min), consisting of 6 binary-response items (that is, yes/no answer), which investigates the following domains: history of falls (that is, the patient has fallen within the last three months), presence of dizziness or lightheadedness, presence of urinary or fecal incontinence, presence of cognitive impairment based on nursing assessment, gait impairment, and mental status (for example presence of any signs of confusion or disorientation), along with impaired judgment or lack of awareness of danger. A total score of 0–1 indicates minimal risk, while a score between 2 and 10 indicates a progressively increasing fall risk, ranging from low to high.

*Falls Efficacy Scale International (FES-I)*: FES-I [40,41] is likely the most extensively utilized questionnaire for assessing fear of falling among elderly adults and individuals with neurological conditions. This questionnaire examines the extent to which an individual is concerned about falling while performing various activities (for example walking on a slippery surface).

#### *2.2.2. Instrumented Assessment*

The ITUG test [28–30] is the instrumental assessment conducted during the TUG test. Specifically, the subjects wore a commercial inertial measurement unit (from mHT-Mhealth technologies, Bologna, Italy) secured to their lower back during the traditional assessment of the TUG test. The validated and commercialized algorithms [29] automatically segmented the TUG test into its phases (that is, sit-to-walk, walking, 180 degrees turn, walk back, turn-to-sit). The IMU recorded 100 distinct measurements, or features, in addition to the total time duration measured by the stopwatch for completing the test. A detailed list of these features is provided in Supplementary Materials. Examples include peak angular velocity during the 180 degrees turn (indicating turning speed) and root mean square of the anterior–posterior acceleration during walking phases (reflecting walking stability). The patients' clinical characteristics are summarized in Table 1.

**Table 1** - Patients' characteristics. TUG, 10 MWT, MB, and FIM—mean (SD) or median (1st to 3rd quartile).

| **Total Number of Patients** | **166** |
|---|---|
| Age, year | 72 (13) |
| Number of elderly over 65 (%) | 125 (73%) |
| Male/female | 97/69 |
| Ischemic/hemorrhagic stroke | 131/35 |
| TUG test, s | 16.55 (9.05) |
| 10 MWT, m/s | 0.96 (0.34) |
| MB, score | 20 (14–23) |
| FIM (total), score | 111 (102–120) |
| FIM motor domain, score | 13.31 (10.94–18.87) |
| FIM cognitive domain, score | 0.96 (0.74–1.17) |

### *2.3. Fall Monitoring*

Falls, defined as "an unintentional coming to rest on the ground or another lower-level surface", were monitored for nine months following the rehabilitation period. Upon discharge, each patient received a calendar to record any falls. Research staff then contacted all participants at the end of the first, second, third, sixth, and ninth months post-discharge to ensure maximum compliance. Based on the number of reported falls, participants were classified into fallers (that is, those who fell at least once) and non-fallers (that is, those who did not fall) during the nine-month follow-up period.

### *2.4. Dataset Composition and Augmentation*

Our study utilizes a subset of the dataset used in [33] and described in Table 1. The dataset includes 142 participants with 108 features in total. Out of 142 participants, data for 15 participants have been synthetically generated to improve the balancing between fallers and non-fallers. Below we provide additional details about the dataset composition and augmentation process.

Firstly, we have limited our analysis exclusively to the following two classes of features:

- Clinical assessment (8 features): It includes the averaged data collected from five repetitions of both the TTD of the TUG tests and the 10 MWT, along with the results from the MB, POMA-B, the Conley Scale, the Falls Efficacy Scale International (FES-I) and the FIM (total score and motor domain).
- ITUG assessment (100 features): Data were obtained from the validated and commercialized ITUG test. A complete list of all features is provided in the Supplementary Materials.

Secondly, we excluded from the original dataset individuals with missing records, such as incomplete discharge assessments or follow-up calls. Excluding participants with missing data was necessary to ensure all features were available for all analyses.

After this step, we obtained a dataset including 127 subjects, out of which only 39 were fallers. To mitigate the imbalance between faller and non-faller patients, in our dataset we augmented the faller group by 38% (that is, 15 individuals). To augment the dataset, we randomly selected 15 fallers and added Gaussian noise to each of their features. The amount of added noise was based on the variance of each respective feature. This process increased the final dataset size to 142 individuals, out of which 54 are fallers and 88 are non-fallers. Next, we divided the dataset into training, validation and test subsets. The training set comprises data from 93 individuals (around 66% of the dataset), with 54 non-fallers and 39 fallers. The validation subset consists of 17 individuals (around 12% of the dataset), with 12 non-fallers and 5 fallers. Finally, the test set is composed of the remaining 32 individuals (22% of the original dataset), with 22 non-fallers subjects and 10 fallers.

Despite the dataset remaining imbalanced, we chose not to further augment the dataset to limit our reliance on synthetic data. The exact amount was determined in order to keep the number of synthetically generated individuals to a minimum, while ensuring at least 10 fallers in test set. Synthetic individuals generated through data augmentation have been excluded from the validation and test sets and have been used exclusively in the training set to avoid introducing any bias in the testing phase.

# 3. Methods

In a preliminary analysis, we trained a Support Vector Machine (SVM) classifier on a balanced subset of 60 patients (30 fallers and 30 randomly sampled non-fallers), using all available features within a 7-fold cross-validation scheme. The resulting performance varies substantially depending on which non-fallers are included in the balanced subset. This instability reflects the intrinsic structure of our dataset, namely a markedly imbalanced population, high inter-subject variability in age and rehabilitation outcomes, and a feature space that is far larger than the number of available subjects.

These characteristics make it unlikely that a single balanced sample or a single training configuration can capture the full variability of the dataset. As a consequence, global discriminative patterns are difficult to identify, and traditional supervised learning approaches are expected to yield unstable results that may not be reliably interpreted. This observation motivates the development of a more robust strategy based on repeated subsampling and stratification, which allows us to explore multiple population configurations and to identify features that consistently distinguish fallers from non-fallers across heterogeneous subsets of the population. This approach categorizes patients into low-, medium-, and high-risk fall groups. To this aim, we implemented a two-step evaluation process. In the first step, we employ a combination of random sampling techniques, SVM classification, and statistical analysis to identify a subset of key features that best distinguish fallers from non-fallers. This procedure results in a set of informative features for fall risk assessment. In the second step, we use these features to develop a risk assessment method for classifying patients into the three fall risk strata (low, medium, or high). The following sections provide a detailed explanation of this approach and of the development of the risk assessment method. The Python implementation of our approach is publicly available[1].

[1] https://github.com/TheEngineRoom-UniGe/RiskOfFallRankingsNotebook

### *3.1. Scale Definition*

This section details our two-step approach to building a fall risk assessment scale. We begin with a comprehensive set of features and use a well-defined process to identify the most informative ones for predicting a specific risk. The process is divided into two parts, namely *feature selection* and *threshold identification*.

#### *3.1.1. Feature Selection*

Given the structure of our dataset, characterized by substantial class imbalance, marked heterogeneity in age, and functional recovery, the identification of globally discriminative patterns proves challenging. Traditional single pass feature selection methods are unable to reliably distinguish fallers from non-fallers under these conditions. For this reason, the IFRA feature selection methodology is deliberately designed around repeated subsampling, which allows us to explore multiple balanced representations of the population without over-relying on any specific configuration of subjects.

We generated 1000 distinct training subsets by keeping the 39 fallers fixed and repeatedly sampling (without replacement) 39 non-fallers out of the available 54. This strategy enables the identification of population subsets in which faller/non-faller separability is sufficiently clear to support robust feature discrimination, while ensuring that no pair of subsets reproduce the same sampling of individuals. For each subset, an SVM classifier is trained, and only subsets achieving more than 80% accuracy on the validation set are retained for further analysis. This criterion functions as a safeguard against overfitting, that is, only population configurations demonstrating stable separability on unseen data contribute to feature selection.

In each retained subset, features are analyzed using a structured statistical procedure via normality testing (Shapiro–Wilk), followed by T-tests or Wilcoxon rank-sum tests as appropriate [42,43]. Features showing statistically significant differences ($p < 0.05$) between fallers and non-fallers are marked as relevant. By aggregating results over all successful subsets, we obtain a consensus ranking of features that consistently distinguish the two groups across heterogeneous population configurations. This consensus-based strategy mitigates overfitting risks inherent to small samples and leverages the inherent variability of the dataset to identify features whose discriminative ability is reproducible across multiple plausible population samplings.

It is important to clarify that in this context the statistical tests (Shapiro–Wilk, T-tests, and Wilcoxon rank-sum tests) are not used with inferential intent, nor are they interpreted as providing clinically generalizable evidence. Rather, they serve as internal discriminators within the subsampling framework, allowing us to identify candidate features that consistently differentiate fallers from non-fallers across heterogeneous balanced subsets of the training population. In datasets characterized by substantial imbalance, high inter-individual variability, and feature dimensionality larger than sample size, such exploratory discriminative procedures are methodologically appropriate and commonly employed to avoid overfitting to any single sample configuration.

The reliability of the selected features stems not from any single test result, but from their repeated emergence across 1000 independent subsampling iterations and under varying population compositions. Only features demonstrating statistically significant separation in a substantial proportion of successful subsamples are retained. This consensus-based selection provides an internal form of reproducibility, that is, the resulting feature set reflects patterns that persist across the intrinsic variability of the dataset, rather than idiosyncrasies of a single partition. Although this does not replace formal external validation, which is an essential next step for future studies, it ensures that the selected features represent the most consistently discriminative signals available within the constraints of the present dataset.

Over 1000 iterations, we keep a list of all the features and track how often each feature is marked as relevant. This list allows us to rank features based on their discriminating power. Features marked as relevant in at least 50% of the iterations are considered highly discriminating and selected for further analysis. A detailed illustration of the process can be found in Figure 1. As the process is independent from the number of input and output features, in the illustration N is the total number of features in the dataset, while M is the number of features deemed relevant by IFRA.

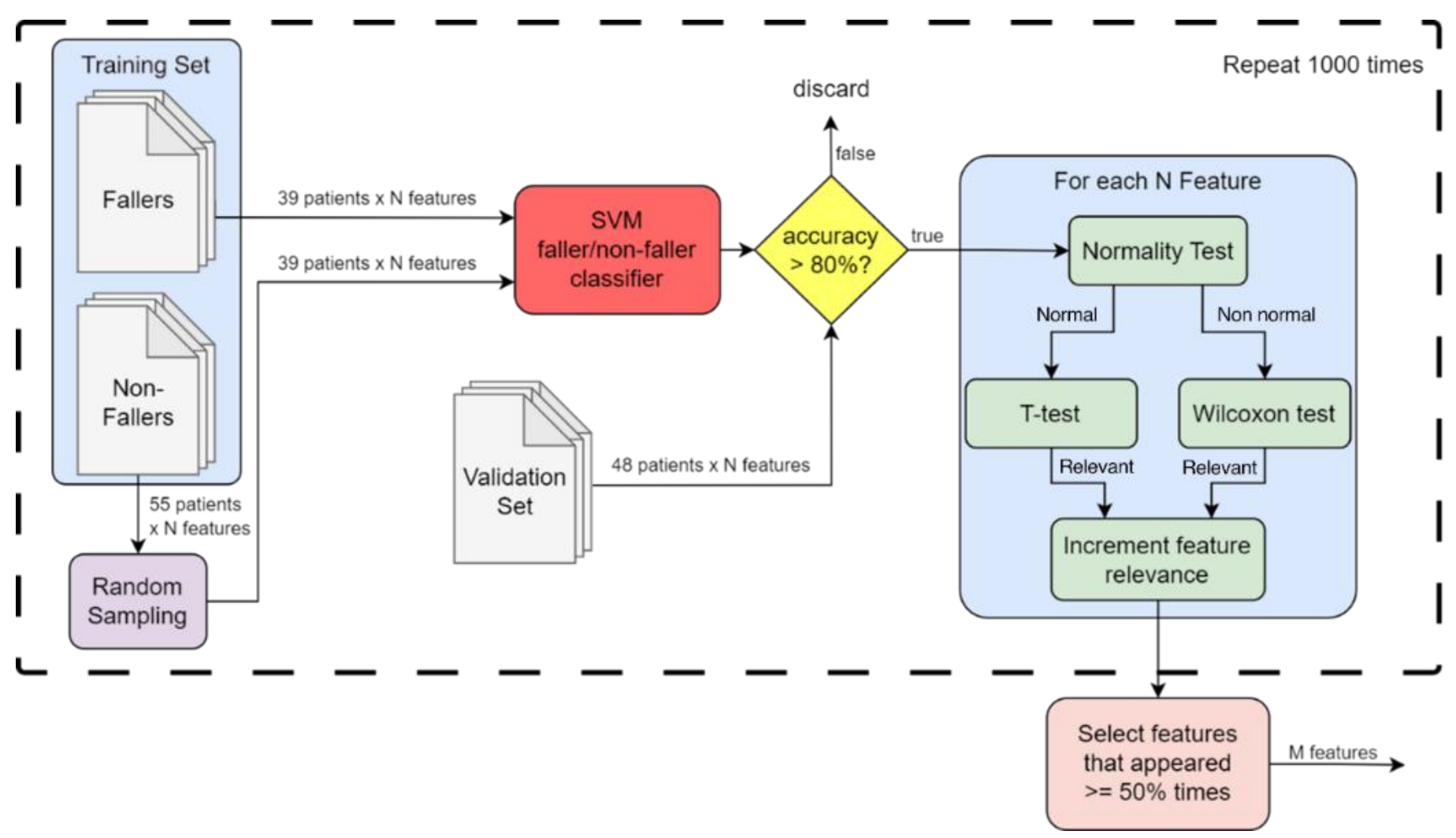

**Figure 1** - A flowchart of the feature selection process used to develop IFRA.

*3.1.2. Thresholds Identification*

The second step in IFRA identifies thresholds to assign new patients to low-, medium-, or high-fall-risk strata using the selected M features. For each feature, we extract and rank the values of all 93 training set subjects. The ranking direction (ascending or descending) depends on the expected relationship with fall risk. For example, features like TUG duration (longer time indicates higher risk) are ranked from lowest to highest, while features like gait speed (faster speed indicates lower risk) are ranked from highest to lowest. Once ranked for a single feature, subjects are divided into three sets of 31 each. These sets represent low-, medium-, and high-risk categories based on that specific feature. The feature values for the subjects at the 31st and 62nd percentiles (delimiting the set in tertiles) are stored for each feature. By iterating through all M features, we establish thresholds based on their distributions within the training data. This allows us to categorize new patients into specific risk strata (low, medium, and high) based on their feature values, ultimately completing the definition of the fall risk assessment scale.

*3.2. Fall Risk Assessment*

Once the fall risk assessment scale is defined, it can be readily applied to classify new patients into risk strata (low, medium, or high). We achieve this by leveraging the percentile values obtained from the training data. The process unfolds as follows:

1. *Feature-based stratification*. For each feature of a new patient, we compare its value to the 31st and 62nd percentile values stored from the training data. Based on this comparison, a preliminary stratum (low, medium, or high) is assigned to the new patient for that specific feature.
2. *Combining Rankings*. Since we have M features, the previous process results in M distinct preliminary strata assignments (one for each feature). To obtain a single fall risk classification, we employ the mode (most frequent value) of these M assignments. In case two strata have the same highest frequency (a tie), we assign the higher risk stratum (medium over low, high over medium).

An interesting aspect of the feature-based stratification process is its similarity to how questionnaires work [44]. In a questionnaire, the numerical value assigned to each answer reflects a monotonic relationship with the underlying quantity being measured (for example, a higher score indicates a higher degree of a specific trait) [35]. Similarly, by assigning patients to tertiles based on their feature values, we can establish a comparable relationship. Each tertile reflects a specific range within the feature's distribution, and these ranges can be considered analogous to answer choices on a questionnaire.

To verify whether the IFRA scale is better than clinical scales at classifying subjects as fallers or non-fallers into the appropriate risk strata, we perform the statistical evaluation described earlier using Fisher's Exact Test. The significance of Fisher's Exact test suggests that there is sufficient statistical evidence to support the association between the risk strata assigned by the IFRA scale and the subject fall risk within this test cohort, thereby supporting the claim that the IFRA scale can be used to estimate fall risk in the target population.

# 4. Results

*4.1. Defining Fall Risk Scales*

Our analyses rely on two distinct feature sets, namely clinical (8 features) and ITUG (100 features). On the ITUG set, we apply the feature selection and risk stratification procedure, described in Section 3, which leads us to identify 22 relevant features and their corresponding threshold values. This process results in the development of IFRA. Features and the corresponding threshold values for IFRA are reported in Table 2.

**Table 2 -** Features list and threshold values defining IFRA. Features are ordered by descending discerning power, based on the number of times they were selected by the feature selection process in Figure 1.

| | **Feature** | **Low Risk Threshold** | **Medium Risk Threshold** | **High Risk Threshold** | **% of Selections** |
|---|---|---|---|---|---|
| 1 | Root Mean Square of the Vertical Acceleration during the Walk Phase [$m/s^2$] | $x \geq 1.91$ | $1.28 < x < 1.91$ | $x \leq 1.28$ | 94% |
| 2 | Range Vertical Acceleration during the Walk Phase [$m/s^2$] | $x \geq 10.58$ | $7.54 < x < 10.58$ | $x \leq 7.54$ | 92% |
| 3 | Root Mean Square of the Angular Velocity about Vertical Axis during the Sit-to-Walk [deg/s] | x ≥ 8.02 | 5.97 < x < 8.02 | x ≤ 5.97 | 86% |
| 4 | Root Mean Square of the Medio-Lateral Acceleration during the Walk Phase [m/s$^2$] | x ≥ 1.28 | 0.99 < x < 1.28 | x ≤ 0.99 | 85% |
| 5 | Range of the Angular Velocity about Vertical Axis during the Sit-to-Walk [deg/s] | x ≥ 32.97 | 24.92 < x < 32.97 | x ≤ 24.92 | 81% |
| 6 | Range Vertical Acceleration during the Sit-to-Walk [m/s$^2$] | x ≥ 5.04 | 3.23 < x < 5.04 | x ≤ 3.23 | 80% |
| 7 | Root Mean Square of the Vertical Acceleration during the Sit-to-Walk [m/s$^2$] | x ≥ 1.23 | 0.88 < x < 1.23 | x ≤ 0.88 | 74% |
| 8 | Gait Speed of ITUG [m/s] | x ≥ 1.13 | 0.72 < x < 1.13 | x ≤ 0.72 | 70% |
| 9 | Peak Angular Velocity of the 180 Turn [deg/s] | x ≥ 120.79 | 89.26 < x < 120.79 | x ≤ 89.26 | 70% |
| 10 | Mean Step Length [m] | x ≥ 0.68 | 0.48 < x < 0.68 | x ≤ 0.48 | 65% |
| 11 | Range Anterior–Posterior Acceleration during the Walk Phase [m/s$^2$] | x ≥ 7.41 | 5.35 < x < 7.41 | x ≤ 5.35 | 63% |
| 12 | Turning Angle of the Turn-to-Sit [deg] | x ≥ 143.55 | 130.87 < x < 143.55 | x ≤ 130.87 | 60% |
| 13 | Peak Angular Velocity of the Turn-to-Sit [deg/s] | x ≥ 142.14 | 94.52 < x < 142.14 | x ≤ 94.52 | 60% |
| 14 | Cadence [steps/min] | x ≥ 109.99 | 92.51 < x < 109.99 | x ≤ 92.51 | 60% |
| 15 | Mean Angular Velocity of the 180 Turn [deg/s] | x ≥ 68.51 | 48.99 < x < 68.51 | x ≤ 48.99 | 58% |
| 16 | Stride Regularity in the Anterior–Posterior Direction [%] | x ≤ 0.39 | 0.39 < x < 0.47 | x ≥ 0.47 | 55% |
| 17 | Normalized Jerk Score in the Anterior–Posterior direction | x ≤ 1.23 | 1.23 < x < 1.6 | x ≥ 1.6 | 55% |
| 18 | Walk/Turn Ratio Return | x ≤ 0.95 | 0.95 < x < 1.3 | x ≥ 1.3 | 52% |
| 19 | Walk Duration [s] | x ≤ 5.54 | 5.54 < x < 8.71 | x ≥ 8.71 | 50% |
| 20 | Walk/Turn Ratio Overall | x ≤ 3.19 | 3.19 < x < 3.9 | x ≥ 3.9 | 50% |
| 21 | Phase Differences Standard Deviation [deg] | x ≤ 11.41 | 11.41 < x < 15.3 | x ≥ 15.3 | 50% |
| 22 | Walk Duration including the 180° Turn [s] | x ≤ 7.89 | 7.89 < x < 11.34 | x ≥ 11.34 | 50% |

We used the clinical features as a benchmark for the classification results obtained by IFRA. All collected clinical features capture relevant elements related to fall risk and have often been used as a potential predictor of falls. While most clinical scales do not stratify patients exactly into three categories, we addressed this by reviewing the literature to identify the optimal threshold values and, in combination with clinical expertise, delineated three meaningful strata for each feature. For instance, the work of Franchignoni et al. [36] considers five strata instead of three for MB scale. To enable a robust comparison with IFRA, we strategically consolidated the three central strata, preserving the extremes. This approach ensured our primary interest to evaluating the accurate assignment of fallers to the uppermost risk strata. It was also found to yield improved results for the MB in our comparison, therefore establishing it as the more conservative option for comparative analysis. The results of this analysis are presented in Table 3, along with references to the articles from which the threshold values were extracted.

**Table 3 -** Features list and corresponding threshold values obtained from literature defining the fall risk scales on clinical features.

| Feature | Low Risk Threshold | Medium Risk Threshold | High Risk Threshold | Reference |
|---|---|---|---|---|
| MB (score) | x ≥ 24.0 | 11.0 < x < 24.0 | x ≤ 11.0 | [36] |
| FIM (total score) | x ≥ 72.0 | 37.0 < x < 72.0 | x ≤ 37.0 | [47] |
| FIM (motor domain, score) | x ≥ 65.0 | 26.0 < x < 65.0 | x ≤ 26.0 | [48] |
| POMA-B (score) | x ≥ 14.0 | 7.0 < x < 14.0 | x ≤ 7.0 | [49] |
| TUG Test (TTD, s) | x ≤ 12.0 | 12.0 < x < 22.0 | x ≥ 22.0 | [50] |
| FES-I (score) | x ≤ 19.0 | 19.0 < x < 28.0 | x ≥ 28.0 | [51] |
| Conley scale (score) | x ≤ 2.0 | 2.0 < x < 7.0 | x ≥ 7.0 | [52] |
| 10 MWT (m/s) | ≥1.0 | 0.6 < x < 1.0 | ≤0.6 | [53] |

*4.2. Fall Risk Evaluation*

To assess the effectiveness of our approach in classifying patients into fall risk strata, we conducted tests on a separate set of 32 unseen volunteers (that is, the test set). These volunteers were purposely excluded from the initial feature selection and scale development phases to avoid bias. It is important to note that the test set exhibits an imbalanced distribution, containing 22 non-fallers and 10 fallers. The results of this evaluation are presented in Table 4 and Figure 2. Table 4 shows the distribution of non-faller (left) and faller (right) individuals across the three strata, based on the classification process using clinical features. Ideally, the desired outcome is for non-fallers to be classified in the low-risk stratum and fallers in the high-risk stratum. It is also of particular importance to highlight that the high-risk category correctly identifies patients who might require extra care after discharge.

Each assessment scale independently ranks patients according to the threshold values extracted from the literature, as detailed before and shown in Table 3. Notably, most scales fail to classify faller patients into the high-risk stratum, with only the FES-I and 10 MWT scales successfully ranking 3 out of 10 faller patients in this category. However, for FES-I, along with other scales such as FIM and Conley, the distribution of fallers and non-fallers across the strata is quite similar. The most effective scales were those based on the MB, POMA-B, and TUG test. To further validate these results, we formulated the following null ($H_0$) and alternative ($H_1$) hypotheses to be tested using Fisher's Exact test [52].

**$H_0$.** *In the population of patients in the test set, the assigned risk strata are not related to their faller or non-faller status*.

**$H_1$.** *In the population of patients in the test set, the assigned risk strata are related to their faller or non-faller status*.

The *p*-values computed using Fisher's Exact Test are reported in the last column of the table. For none of the considered scales was it possible to reject the null hypothesis ($p < 0.05$), therefore suggesting a weak relationship between the assigned strata and fall risk. Nevertheless, the limited number of subjects and the imbalance between fallers and non-fallers could have affected the results.

**Table 4** - Classification results on the test set using the clinical features, with threshold values from the literature.

| Feature | Non-Fallers | | | Fallers | | | |
|---|---|---|---|---|---|---|---|
| | Low | Medium | High | Low | Medium | High | *p*-Value |
| MB | 27.3% | 72.7% | 0.0% | 10.0% | 70.0% | 20.0% | 0.119 |
| FIM (total) | 95.4% | 4.6% | 0.0% | 90.0% | 10.0% | 0.0% | 0.534 |
| FIM (motor domain) | 95.4% | 4.6% | 0.0% | 80.0% | 20.0% | 0.0% | 0.224 |
| POMA-B | 59.1% | 40.9% | 0.0% | 50.0% | 40.0% | 10.0% | 0.228 |
| TUG Test (TTD) | 45.4% | 45.4% | 9.2% | 20.0% | 60.0% | 20.0% | 0.379 |
| FES-I | 31.8% | 40.9% | 27.3% | 20.0% | 50.0% | 30.0% | 0.890 |
| Conley Scale | 72.7% | 27.3% | 0.0% | 60.0% | 30.0% | 10.0% | 0.454 |
| 10 MWT | 68.1% | 27.3% | 4.6% | 50.0% | 20.0% | 30.0% | 0.625 |

In contrast, Figure 2 presents the results of the same classification performed using the IFRA scale (Table 2). In this case, most fallers (6 out of 10) are successfully categorized as high risk. A comparison between Table 4 and Figure 2 also reveals a noticeable increase in the number of fallers classified as high risk when using the IFRA scale, which leverages features from the ITUG. As hinted in Section 4.2, that is a desirable behavior as it allows for better identifying patients who might require extra care after discharge. Furthermore, the IFRA scale performs consistently better also on non-fallers compared to the clinical scales. Most non-fallers are classified as low risk, with only 2 subjects out of 22 classified as high risk.

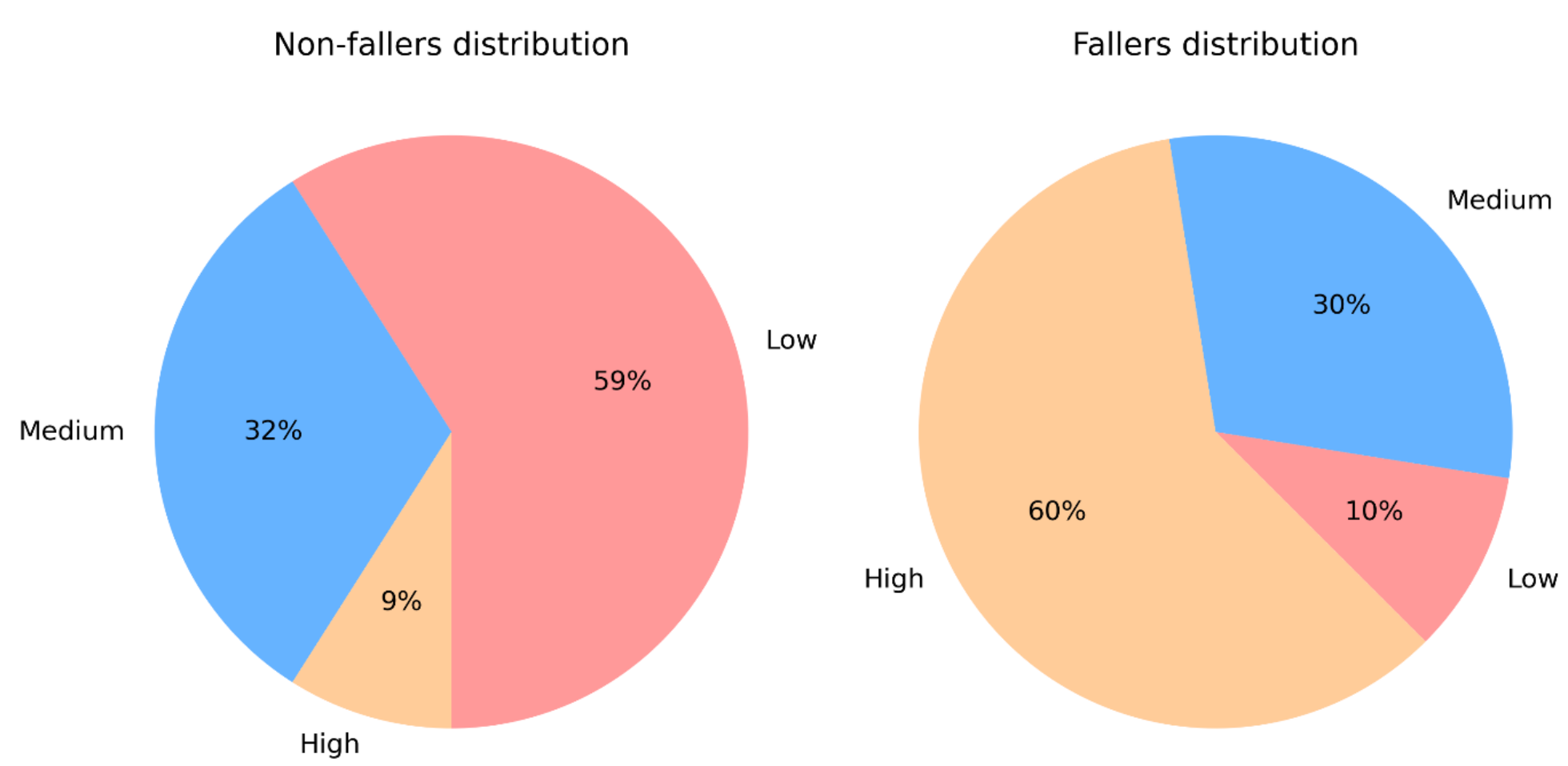

**Figure 2** - Distribution of faller and non-faller individuals in the test set according to IFRA.

At first glance, the IFRA scale appears to perform favorably in classifying subjects into risk strata aligned with their faller or non-faller status. Fisher's Exact test yields a *p*-value of 0.004, suggesting a statistically detectable association between IFRA-derived strata and fall status within this specific test cohort. However, this result must be interpreted in the context of the small and imbalanced test set (10 fallers) and therefore should not be viewed as definitive evidence of generalizable performance. Rather, it supports the notion that IFRA may capture meaningful mobility patterns within the constraints of the present dataset, motivating further validation in larger and more diverse populations.

# 5. Discussion

*5.1. Analysis of the Results*

In this study, we investigate a new proposal for a fall risk index aimed at stroke patients using machine learning, which we call IFRA, to categorize patients' fall risk.

In a preliminary phase, we identify the features that predispose individuals to a higher risk of falling, including both clinical measures derived from routine clinical practice and instrumental measures collected during an ITUG test. Based on these results, IFRA is designed and developed. While this tool is constructed using machine learning methods, it ultimately results in a table of significant features with threshold values that clinicians can use to assign a patient to one of three risk strata. Similarly to widely used clinical evaluation scales, this can be performed by collecting measurements of the significant features for a given patient and then using the corresponding threshold values to assign each feature to a risk stratum. The patient's overall risk stratum is then determined by the mode of the features' risk-strata assignments. It follows that all features are considered to have equal importance within the IFRA scale. Table 2 not only reports threshold values for the low-, medium-, and high-risk strata, but also orders the selected features according to their descending discriminative power.

The features ranked at the top of the IFRA list are consistent with findings previously reported in the literature on neurological populations. Specifically, vertical acceleration and angular velocity during walking are expected to exhibit lower root mean square values in older adults with a higher risk of falling [53] and in stroke patients with reduced mobility [54]. In addition, mediolateral acceleration during walking indexes lateral instability, as reductions in mediolateral control are known to underlie poor balance and increased fall risk [28]. Similarly, during the sit-to-walk transition, reductions in vertical acceleration and angular velocity have been associated with impaired dynamic balance in individuals with Parkinson's disease [32]. Peak angular velocity during turning emerges as a particularly sensitive marker of motor impairment. Indeed, slower turning velocities have been shown to correlate with poorer functional mobility in individuals with chronic stroke [55], reduced balance performance [30,31], and disease-specific gait deficits in Parkinson's disease [56]. Further supporting the clinical relevance of turning-related metrics, turning duration during the ITUG test and the MB have demonstrated substantial criterion validity in predicting the likelihood of future falls within a nine-month follow-up period in neurological patients [33].

When we perform additional analyses considering only clinical variables (that is, excluding instrumental ones), results show that, when instrumental measurements cannot be obtained due to clinical constraints, traditional clinical scales can still provide useful insights into a patient's fall risk.

Within the constraints of our dataset, the MB scale emerged as the clinical measure most closely aligned with fall status. This observation is consistent with findings from our previous study [33], although such consistency should not be interpreted as confirming broad predictive validity. In particular, the apparent ability of MB to capture balance-related aspects of fall risk may be influenced by sample characteristics, rehabilitation context, and the limited number of fall events available for analysis.

The traditional TTD, which is frequently employed as an indicator of dynamic balance [30], also performed comparatively well, as expected for a measure widely used to assess functional mobility. Nonetheless, several ITUG-derived features appeared to capture mobility subtleties not reflected by TTD alone. These findings should be regarded as preliminary indications rather than evidence of superiority; that is, they suggest that instrumented assessments may offer complementary insights when clinical conditions allow their adoption, but further research is required to determine whether such advantages persist in larger and more diverse cohorts.

### *5.2. Future Perspectives*

Recent advancements in wearable technology, especially when using a single IMU, as in the case of the ITUG adopted in this study, together with improvements in computational platforms for motion data analysis, have significantly broadened the available options for monitoring and evaluating patients in a variety of settings.

In fact, over the last decades, IMUs have been increasingly used in movement analysis and assessment, including in patients affected by neurological conditions [33]. IMUs have transformed the way human movement is recorded by leveraging their small size, extremely low weight, relatively low cost, and ease of use. Their adoption is now widespread in clinical practice and not limited to research settings [15,57].

According to the findings of this study, instrumental motion analysis demonstrates a high capability to identify individuals at risk of falling, comparable to that of clinical measurements alone. This result may support the selection of appropriate measures for defining fall risk according to the specific context of the patient, whether in a hospital or in a home-based setting.

Obtaining accurate and timely patient information, particularly from individuals with impaired mobility, is a critical issue that requires careful consideration in future studies. This, in turn, enables frequent and straightforward monitoring of physical performance and fall risk, for instance at home and without the need for clinical personnel.

Incorporating remote monitoring of gait and balance into post-discharge home care may further enhance fall-risk identification by revealing mobility impairments that are not evident in controlled clinical environments [58,59]. In principle, IFRA enables the timely detection of motor decline that may lead to an increased fall risk, prompting patients or their caregivers to seek medical attention at an earlier stage. Therefore, monitoring activities may also serve a preventive role with respect to future fall events.

Patients could utilize IMUs during exercise routines involving sit-to-walk transfers and walking along curved paths. Information about anomalies in vertical acceleration or angular velocity during these tasks can be readily assessed and conveyed to patients, prompting adjustments in performance aimed at improving these parameters. In this context, automated analysis of IMU signals through dedicated algorithms and models is essential to provide actionable feedback. In addition, it enables the examination of within-person changes over time, which can improve the discrimination capabilities of predictive models [60].

However, the success of any remote patient monitoring system largely depends on the practicality and usability of the technology from the patient’s perspective. Numerous tools are available for remotely assessing gait and balance, including external sensors (for example, cameras and force plates) and IMUs. Unlike external sensors, which may require expensive hardware and dedicated installation, IMUs are portable, inexpensive, and—when considering smartphones, which are inherently equipped with IMUs—nearly ubiquitous [28,61]. Describing and understanding the drivers of participant engagement with remote patient monitoring will be necessary to determine the success of future real-world implementations. The primary challenges associated with remote patient monitoring include system usability, data quality, safety during at-home functional tasks, and barriers to long-term adherence [62,63].

### *5.3. Limitations*

In this article, we propose a novel methodology to derive a fall risk assessment scale based on ITUG data. While the preliminary results presented in the previous sections compare rather favorably with traditional clinical evaluation scales, some limitations of this study must be highlighted, namely (i) the reduced sample size; (ii) the use of algorithms to automatically segment the ITUG test into its constituent phases; (iii) the absence of solid normative data references in the literature for a healthy age-matched population; and (iv) the lack, within the dataset, of additional information that could support fall risk assessment, such as trunk trajectory data and detailed step characteristics.

Sample size (i) is undoubtedly the primary limitation of this work. However, the number of recruited patients is in line with that of other studies [64,65], and the findings obtained offer promising indications of the proposed method’s potential for fall risk assessment. To fully validate the efficacy and applicability of the proposed method, further investigation is required, particularly through larger-scale studies and testing across different pathologies (such as Parkinson’s disease).

The use of algorithms to automatically segment the ITUG test into its constituent phases (ii) represents another potential limitation, as these phases are subsequently used to derive additional measures employed in our analysis. While the reliability of these segmentation algorithms has been previously demonstrated [29], it is important to note that automatic methods may fail in certain conditions [66], especially in populations with more severe motor impairment. For example, as gait and mobility become increasingly pathological (e.g., very high TTD values), segmentation algorithms may struggle to correctly identify mobility patterns and phase boundaries within the TUG test. Nevertheless, results obtained on the test set remain encouraging, at least for the population considered in this study. Further investigations are therefore needed to assess both the impact of different segmentation algorithms and the applicability of automatic segmentation techniques in patients with higher levels of motor impairment.

The absence of solid normative data references in the literature for a healthy age-matched population (iii) is another limitation of this work, which directly reflects the current state of the art. While the definition of such normative references lies outside the scope of the present study, future investigations aimed at extending the applicability of IFRA should address this issue.

Finally, the lack of additional information in the dataset that could further support fall risk assessment (iv) represents another area for improvement. In this work, we employed the data collected in [33] and showed that the available features are sufficient to provide promising indications of IFRA's potential for assessing fall risk. Nevertheless, future data collection efforts may include additional potentially informative features, such as trunk trajectory data and detailed step-related measures. The feature selection methodology proposed in this study and used to derive the IFRA scale could then be seamlessly applied to evaluate the contribution of such features to the assessment framework.

A further limitation concerns the exploratory nature of the feature selection methodology. Although repeated subsampling and statistical discrimination allow us to identify features that are consistently associated with fall risk across heterogeneous subsets of the population, this approach does not, by itself, establish external validity or formal reproducibility of the IFRA scale. The statistical tests employed within the pipeline are intended solely as internal discriminators for feature selection and should not be interpreted as providing inferential evidence at the population level. Moreover, while the emergence of stable feature patterns across 1000 subsampling iterations supports the internal robustness of the methodology, confirming the reproducibility of IFRA in independent cohorts will require future studies involving larger and more representative datasets.

It is therefore essential to consider IFRA not as a fully validated clinical tool, but rather as a methodological proposal tailored to highly imbalanced and heterogeneous data environments, such as those typically encountered in post-rehabilitation cohorts. The three-strata scale derived from this methodology is designed to complement—rather than replace—traditional clinical assessments, providing a framework that can be integrated alongside established measures. Further research will be required to evaluate the scale's performance, generalizability, and clinical utility across broader populations and settings.

# 6. Conclusions

In this study, we explore the potential of IMUs coupled with machine learning techniques to advance fall risk assessments in post-stroke rehabilitation. The motivation behind this research is to leverage the capabilities of IMUs to transform the traditional TUG test into a more nuanced source of mobility measures.

This study introduces a proof-of-concept methodology for deriving a fall risk assessment scale, referred to as the Instrumented Fall Risk Assessment (IFRA), based on ITUG-derived features. The findings highlight signals that warrant further investigation; however, they should not be interpreted as establishing IFRA's clinical validity or superiority over established scales. Rather, the present results serve as an initial demonstration that an automated, feature-based stratification approach may complement traditional assessments by capturing aspects of mobility that merit additional study.

Our method involves feature selection using a Support Vector Machine classifier to identify significant features and corresponding thresholds across randomized subsets of training data. The discriminative features identified—such as gait speed measured during the ITUG test, vertical acceleration during the sit-to-walk transition, and turning angular velocity—are consistent with findings reported in existing studies and with our prior investigations on balance-related measures. Moreover, the identified thresholds enable stratification of patients into specific fall risk levels based on features extracted from the ITUG. Overall, our results suggest that IFRA may be compared with traditional clinical scales in the context of fall risk identification among stroke patients.

As noted above, despite these encouraging findings, this study presents limitations in terms of generalizability, primarily due to the relatively small dataset size. Future work involving larger and more representative cohorts will be essential to evaluate the generalizability, reliability, and clinical utility of IFRA.

**Author Contributions:** S.M.: methodology, software, formal analysis, validation, and conceptualization. A.C. (Alessandro Carfì): writing (review & editing), methodology, supervision, and conceptualization. A.C. (Alessio Capitanelli): project administration, writing (original draft), methodology, supervision, and conceptualization. P.T.: writing (review & editing) and conceptualization. M.C.: supervision and conceptualization. F.M.: writing (review & editing), supervision, and conceptualization. M.P.: writing (review & editing), data curation, supervision, and conceptualization. All authors have read and agreed to the published version of the manuscript.

**Funding:** This work was also funded by the European Union (NextGenerationEU); by the Ministry of University and Research (MUR), National Recovery and Resilience Plan (NRRP), Mission 4, Component 2, Investment 1.5, project "RAISE—Robotics and AI for Socio-economic Empowerment" (ECS00000035); and by the Italian Ministry of Research, under the complementary actions to the National Recovery and Resilience Plan (NRRP) "Fit4MedRob-Fit for Medical Robotics" Grant (# PNC0000007).

**Institutional Review Board Statement:** The studies involving humans were approved by Comitato Etico Milano Area 2, Milan, Italy (568_2018bis; 25 October 2018). The studies were conducted in accordance with the local legislation and institutional requirements. The participants provided their written informed consent to participate in this study.

**Data Availability Statement:** The dataset supporting the conclusions of this article will be made available by the authors, without undue reservation.

**Conflicts of Interest:** Authors Simone Macciò and Alessio Capitanelli was employed by the company Teseo Srl. The remaining authors declare that the research was conducted in the absence of any commercial or financial relationships that could be construed as a potential conflict of interest.

**References**

1. Chandran, A.; Hyder, A.A.; Peek-Asa, C. The global burden of unintentional injuries and an agenda for progress. *Epidemiol. Rev.* **2010**, *32*, 110–120.

2. Langhorne, P.; Stott, D.; Robertson, L.; MacDonald, J.; Jones, L.; McAlpine, C.; Dick, F.; Taylor, G.; Murray, G. Medical complications after stroke: A multicenter study. *Stroke* **2000**, *31*, 1223–1229.

3. Weerdesteijn, V.; Niet, M.D.; Van Duijnhoven, H.; Geurts, A.C. Falls in individuals with stroke. *J. Rehabil. Res. Dev.* **2008**, *45*, 1195–1214.

4. Melillo, P.; Orrico, A.; Scala, P.; Crispino, F.; Pecchia, L. Cloud-based smart health monitoring system for automatic cardiovascular and fall risk assessment in hypertensive patients. *J. Med. Syst.* **2015**, *39*, 109.

5. Wei, T.-S.; Liu, P.-T.; Chang, L.-W.; Liu, S.-Y. Gait asymmetry, ankle spasticity, and depression as independent predictors of falls in ambulatory stroke patients. *PLoS ONE* **2017**, *12*, e0177136.

6. Mackintosh, S.F.; Hill, K.D.; Dodd, K.J.; Goldie, P.A.; Culham, E.G. Balance score and a history of falls in hospital predict recurrent falls in the 6 months following stroke rehabilitation. *Arch. Phys. Med. Rehabil.* **2006**, *87*, 1583–1589.

7. Holloway, R.G.; Tuttle, D.; Baird, T.; Skelton, W.K. The safety of hospital stroke care. *Neurology* **2007**, *68*, 550–555.

8. Dennis, M.; Lo, K.; McDowall, M.; West, T. Fractures after stroke: Frequency, types, and associations. *Stroke* **2002**, *33*, 728–734.

9. Pouwels, S.; Lalmohamed, A.; Leufkens, B.; de Boer, A.; Cooper, C.; van Staa, T.; de Vries, F. Risk of hip/femur fracture after stroke: A population-based case-control study. *Stroke* **2009**, *40*, 3281–3285.

10. Caronni, A.; Picardi, M.; Redaelli, V.; Antoniotti, P.; Pintavalle, G.; Aristidou, E.; Gilardone, G.; Carpinella, I.; Lencioni, T.; Arcuri, P.; et al. The Falls Efficacy Scale International is a valid measure to assess the concern about falling and its changes induced by treatments. *Clin. Rehabil.* **2022**, *36*, 558–570.

11. Walsh, M.E.; Horgan, N.F.; Walsh, C.D.; Galvin, R. Systematic review of risk prediction models for falls after stroke. *J. Epidemiol. Community Health* **2016**, *70*, 513–519.

12. Tan, K.M.; Tan, M.P. Stroke and falls—Clash of the two titans in geriatrics. *Geriatrics* **2016**, *1*, 31.

13. Xu, T.; Clemson, L.; O'Loughlin, K.; Lannin, N.A.; Dean, C.; Koh, G. Risk factors for falls in community stroke survivors: A systematic review and meta-analysis. *Arch. Phys. Med. Rehabil.* **2018**, *99*, 563–573.

14. Rogers, M.E.; Rogers, N.L.; Takeshima, N.; Islam, M.M. Methods to assess and improve the physical parameters associated with fall risk in older adults. *Prev. Med.* **2003**, *36*, 255–264.

15. Davis, E.; Periassamy, M.; Stock, B.; Altenburger, P.; Ambike, S.; Haddad, J. Using inertial measurement units (IMUs) to detect mobility declines in middle-aged individuals. *Innov. Aging* **2023**, *7*, 671.

16. Alexander, B.H.; Rivara, F.P.; Wolf, M.E. The cost and frequency of hospitalization for fall-related injuries in older adult. *Am. J. Public Health* **1992**, *82*, 1020–1023.

17. American Geriatrics Society; British Geriatrics Society; American Academy of Orthopaedic Surgeons Panel on Fall Prevention. Guideline for the prevention of falls in older persons. *J. Am. Geriatr. Soc.* **2001**, *49*, 664–672.

18. Panel on Prevention of Falls in Older Persons. American geriatrics society and British geriatrics society: Summary of the updated American geriatrics Society/British geriatrics society clinical practice guideline for prevention of falls in older persons. *J. Am. Geriatr. Soc.* **2011**, *59*, 148–157.

19. Shumway-Cook, A.; Brauer, S.; Woollacott, M. Predicting the probability for falls in community-dwelling older adults using the Timed Up& Go Test. *Phy. Ther.* **2000**, *80*, 896.

20. Schoene, D.; Wu, S.M.-S.; Mikolaizak, A.S.; Menant, J.C.; Smith, S.T.; Delbaere, K.; Lord, S.R. Discriminative ability and predictive validity of the timed Up and Go test in identifying older people who fall: Systematic review and meta-analysis. *J. Am. Geriatr. Soc.* **2013**, *61*, 202–208.

21. Podsiadlo, D.; Richardson, S. The timed Up& Go: A test of basic functional mobility for frail elderly persons. *J. Am. Geriatr. Soc.* **1991**, *39*, 142–148.

22. Wu, X.; Yeoh, H.T.; Lockhart, T. Fall risks assessment and fall prediction among community dwelling elderly using wearable wireless sensors. *Proc. Hum. Factors Ergon. Soc. Annu. Meet.* **2013**, *57*, 109–113.

23. Viccaro, L.J.; Perera, S.; Studenski, S.A. Is timed up and go better than gait speed in predicting health, function, and falls in older adults? *J. Am. Geriatr. Soc.* **2011**, *59*, 887–892.

24. Okumiya, K.; Matsubayashi, K.; Nakamura, T.; Fujisawa, M.; Osaki, Y.; Doi, Y.; Ozawa, T. The Timed Up& Go test is a useful predictor of falls in community-dwelling older people. *J. Am. Geriatr. Soc.* **1998**, *46*, 928–929.

25. Salarian, A.; Horak, F.B.; Zampieri, C.; Carlson-Kuhta, P.; Nutt, J.G.; Aminian, K. iTUG, a sensitive and reliable measure of mobility. *IEEE Trans. Neural Syst. Rehabil. Eng.* **2010**, *18*, 303–310.

26. Patel, M.; Pavic, A.; Goodwin, V.A. Wearable inertial sensors to measure gait and posture characteristic differences in older adult fallers and non-fallers: A scoping review. *Gait Posture* **2020**, *76*, 110–121.

27. Petraglia, F.; Scarcella, L.; Pedrazzi, G.; Brancato, L.; Puers, R.; Costantino, C. Inertial sensors versus standard systems in gait analysis: A systematic review and meta-analysis. *Eur. J. Phys. Rehabil. Med.* **2019**, *55*, 265–280.

28. Ortega-Bastidas, P.; Gomez, B.; Aqueveque, P.; Luarte-Martínez, S.; Cano-de-la-Cuerda, R. Instrumented Timed Up and Go Test (ITUG)-more than assessing time to predict falls: A systematic review. *Sensors* **2023**, *23*, 3426.

29. Mellone, S.; Tacconi, C.; Chiari, L. Validity of a Smartphone-based instrumented Timed Up and Go. *Gait Posture* **2012**, *36*, 163–165.

30. Caronni, A.; Sterpi, I.; Antoniotti, P.; Aristidou, E.; Nicolaci, F.; Picardi, M.; Pintavalle, G.; Redaelli, V.; Achille, G.; Sciumè, L.; et al. Criterion validity of the instrumented Timed Up and Go test: A partial least square regression study. *Gait Posture* **2018**, *61*, 287–293.

31. Caronni, A.; Picardi, M.; Aristidou, E.; Antoniotti, P.; Pintavalle, G.; Redaelli, V.; Sterpi, I.; Corbo, M. How do patients improve their timed up and go test? Responsiveness to rehabilitation of the TUG test in elderly neurological patients. *Gait Posture* **2019**, *70*, 33–38.

32. Picardi, M.; Redaelli, V.; Antoniotti, P.; Pintavalle, G.; Aristidou, E.; Sterpi, I.; Meloni, M.; Corbo, M.; Caronni, A. Turning and sit-to-walk measures from the instrumented Timed Up and Go test return valid and responsive measures of dynamic balance in Parkinson's disease. *Clin. Biomech.* **2020**, *80*, 105177.

33. Caronni, A.; Picardi, M.; Scarano, S.; Malloggi, C.; Tropea, P.; Gilardone, G.; Aristidou, E.; Pintavalle, G.; Redaelli, V.; Antoniotti, P.; et al. Pay attention: You can fall! The Mini-BESTest scale and the turning duration of the TUG test provide valid balance measures in neurological patients: A prospective study with falls as the balance criterion. *Front. Neurol.* **2023**, *14*, 1228302.

34. Studenski, S.; Perera, S.; Patel, K.; Rosano, C.; Faulkner, K.; Inzitari, M.; Brach, J.; Chandler, J.; Cawthon, P.; Connor, E.B.; et al. Gait speed and survival in older adults. *JAMA* **2011**, *305*, 50–58. https://doi.org/10.1001/jama.2010.1923. PMID: 21205966; PMCID: PMC3080184.

35. Tesio, L.; Scarano, S.; Hassan, S.; Kumbhare, D.; Caronni, A. Why Questionnaire Scores are not Measures: A question-raising article. *Am. J. Phys. Med. Rehabil.* **2023**, *102*, 75.

36. Franchignoni, F.; Horak, F.; Godi, M.; Nardone, A.; Giordano, A. Using psychometric techniques to improve the Balance Evaluation System's Test: The mini-BESTest. Journal of rehabilitation medicine: Official journal of the UEMS European Board of Physical and Rehabilitation *Medicine* **2010**, *42*, 323.

37. Caronni, A.; Picardi, M.; Scarano, S.; Tropea, P.; Gilardone, G.; Bolognini, N.; Redaelli, V.; Pintavalle, G.; Aristidou, E.; Antoniotti, P.; et al. Differential item functioning of the mini-BESTest balance measure: A Rasch analysis study. *Int. J. Environ. Res. Public Health* **2023**, *20*, 5166.

38. Linacre, J.M.; Heinemann, A.W.; Wright, B.D.; Granger, C.V.; Hamilton, B.B. The structure and stability of the Functional Independence Measure. *Arch. Phys. Med. Rehabil.* **1994**, *75*, 127–132.

39. Tinetti, M.E. Performance-oriented assessment of mobility problems in elderly patients. *J. Am. Geriatr. Soc.* **1986**, *34*, 119–126.

40. Conley, D.; Schultz, A.A.; Selvin, R. The challenge of predicting patients at risk for falling: Development of the Conley Scale. *Medsurg Nurs.* **1999**, *8*, 348–354.

41. Yardley, L.; Beyer, N.; Hauer, K.; Kempen, G.; Piot-Ziegler, C.; Todd, C. Development and initial validation of the falls efficacy scale-international (FES-I). *Age Ageing* **2005**, *34*, 614–619.

42. Wilcoxon, F. Individual comparisons by ranking methods. In *Breakthroughs in Statistics: Methodology and Distribution*; Springer: New York, NY, USA, 1992; pp. 196–202.

43. Wilcoxon, F. Probability tables for individual comparisons by ranking methods. *Biometrics* **1947**, *3*, 119–122.

44. Mathias, S.; Nayak, U.S.; Isaacs, B. Balance in elderly patients: The "get-up and go" test. *Arch. Phys. Med. Rehabil.* **1986**, *67*, 387–389.

45. Inouye, M.; Hashimoto, H.; Mio, T.; Sumino, K. Influence of admission functional status on functional change after stroke rehabilitation. *Am. J. Phys. Med. Rehabil.* **2001**, *80*, 121–125.

46. Kwon, S.; Hartzema, A.G.; Duncan, P.W.; Min-Lai, S. Disability measures in stroke: Relationship among the Barthel Index, the Functional Independence Measure, and the Modified Rankin Scale. *Stroke* **2004**, *35*, 918–923.

47. Verghese, J.; Buschke, H.; Viola, L.; Katz, M.; Hall, C.; Kuslansky, G.; Lipton, R. Validity of divided attention tasks in predicting falls in older individuals: A preliminary study. *J. Am. Geriatr. Soc.* **2002**, *50*, 1572–1576.

48. Vieira, E.R.; Palmer, R.C.; Chaves, P.H. Prevention of falls in older people living in the community. *BMJ* **2016**, *353*, i1419.

49. Delbaere, K.; Close, J.C.; Mikolaizak, A.S.; Sachdev, P.S.; Brodaty, H.; Lord, S.R. The falls efficacy scale international (FES-I). A comprehensive longitudinal validation study. *Age Ageing* **2010**, *39*, 210–216.

50. Lovallo, C.; Rolandi, S.; Rossetti, A.M.; Lusignani, M. Accidental falls in hospital inpatients: Evaluation of sensitivity and specificity of two risk assessment tools. *J. Adv. Nurs.* **2010**, *66*, 690–696.

51. Quach, L.; Galica, A.M.; Jones, R.N.; Procter-Gray, E.; Manor, B.; Hannan, M.T.; Lipsitz, L.A. The nonlinear relationship between gait speed and falls: The maintenance of balance, independent living, intellect, and zest in the elderly of Boston study. *J. Am. Geriatr. Soc.* **2011**, *59*, 1069–1073.

52. Fisher, R.A. On the interpretation of χ 2 from contingency tables, and the calculation of P. *J. R. Stat. Soc.* **1922**, *85*, 87–94.

53. Zakaria, N.A.; Kuwae, Y.; Tamura, T.; Minato, K.; Kanaya, S. Quantitative analysis of fall risk using TUG test. *Comput. Methods Biomech. Biomed. Eng.* **2015**, *18*, 426–437.

54. Higashi, Y.; Yamakoshi, K.; Fujimoto, T.; Sekine, M.; Tamura, T. Quantitative evaluation of movement using the timed up-and-go test. *IEEE Eng. Med. Biol. Mag.* **2008**, *27*, 38–46.

55. Spina, S.; Facciorusso, S.; D'ASCANIO, M.C.; Morone, G.; Baricich, A.; Fiore, P.; Santamato, A. Sensor based assessment of turning during instrumented Timed Up and Go Test for quantifying mobility in chronic stroke patients. *Eur. J. Phys. Rehabil. Med.* **2022**, *59*, 6.

56. Zampieri, C.; Salarian, A.; Carlson-Kuhta, P.; Aminian, K.; Nutt, J.G.; Horak, F.B. The instrumented timed up and go test: Potential outcome measure for disease modifying therapies in Parkinson's disease. *J. Neurol. Neurosurg. Psychiatry* **2010**, *81*, 171–176.

57. Voisard, C.; de L'escalopier, N.; Ricard, D.; Oudre, L. Automatic gait events detection with inertial measurement units: Healthy subjects and moderate to severe impaired patients. *J. Neuroeng. Rehabil.* **2024**, *21*, 104. https://doi.org/10.1186/s12984-024-01405-x.

58. Weiss, A.; Brozgol, M.; Dorfman, M.; Herman, T.; Shema, S.; Giladi, N.; Hausdorff, J.M. Does the evaluation of gait quality during daily life provide insight into fall risk? A novel approach using 3-day accelerometer recordings. *Neurorehabil. Neural Repair.* **2013**, *27*, 742–752.

59. Suffoletto, B.; Kim, D.; Toth, C.; Mayer, W.; Glaister, S.; Cinkowski, C.; Ashenburg, N.; Lin, M.; Losak, M. Feasibility of Measuring Smartphone Accelerometry Data During a Weekly Instrumented Timed Up-and-Go Test After Emergency Department Discharge: Prospective Observational Cohort Study. *JMIR Aging* **2024**, *7*, e57601.

60. Mancini, M.; Horak, F.B. Potential of APDM mobility lab for the monitoring of the progression of Parkinson's disease. *Expert Rev. Med. Devices.* **2016**, *13*, 455–462.

61. Manor, B.; Yu, W.; Zhu, H.; Harrison, R.; Lo, O.-Y.; Lipsitz, L.; Travison, T.; Pascual-Leone, A.; Zhou, J. Smartphone app-based assessment of gait during normal and dual-task walking: Demonstration of validity and reliability. *JMIR mHealth uHealth* **2018**, *6*, e36.

62. Kobsar, D.; Charlton, J.M.; Tse, C.T.F.; Esculier, J.-F.; Graffos, A.; Krowchuk, N.M.; Thatcher, D.; Hunt, M.A. Validity and reliability of wearable inertial sensors in healthy adult walking: A systematic review and meta-analysis. *J. Neuroeng. Rehabil.* **2020**, *17*, 62.

63. Kristoffersson, A.; Du, J.; Ehn, M. Performance and characteristics of wearable sensor systems discriminating and classifying older adults according to fall risk: A systematic review. *Sensors* **2021**, *21*, 5863.

64. Minet, L.R.; Peterson, E.; Von Koch, L.; Ytterberg, C. Occurrence and predictors of falls in people with stroke: Six-year prospective study. *Stroke* **2015**, *46*, 2688–2690.

65. Lim, Z.K.; Connie, T.; Ong Michael Goh, K.; Saedon, N.I. Fall Risk Prediction Using Temporal Gait Features and Machine Learning Approaches. *Front. Artif. Intell.* **2024**, *7*, 1425713.

66. Chubak, J.; Pocobelli, G.; Weiss, N.S. Tradeoffs between accuracy measures for electronic health care data algorithms. *J. Clin. Epidemiol.* **2012**, *65*, 343–349.

# A - Supplementary Materials

## *A.1 List of features extrapolated from data acquired through IMU*

**Table A.1** - Full list of features obtained from processing the inertial data, split according to the constituent phases of the TUG test. For each feature, the corresponding unit of measurement is reported in square brackets, if applicable.

| Instrumented TUG Features |
|---|
| Total Duration [s] |
| Sit-to-Walk Duration [s] |
| 180° Turn Duration [s] |
| Turn Duration in the Turn-to-Sit Phase [s] |
| Turn-to-Sit Duration [s] |
| Walk/Turn Ratio Outward |
| Walk/Turn Ratio Return |
| Walk/Turn Ratio Overall |
| Walk Duration including the 180° Turn [s] |
| Total Number of Steps |
| Range Anterior-Posterior Acceleration during the Sit-to-Walk Transition [m/s^2] |
| Range Medio-Lateral Acceleration during the Sit-to-Walk Transition [m/s^2] |
| Range Vertical Acceleration during the Sit-to-Walk Transition [m/s^2] |
| Root Mean Square of the Anterior-Posterior Acceleration during the Sit-to-Walk Transition [m/s^2] |
| Root Mean Square of the Medio-Lateral Acceleration during the Sit-to-Walk Transition [m/s^2] |
| Root Mean Square of the Vertical Acceleration during the Sit-to-Walk Transition [m/s^2] |
| Jerk Score Anterior-Posterior Acceleration during the Sit-to-Walk Transition [m] |
| Jerk Score Medio-Lateral Acceleration during the Sit-to-Walk Transition [m] |
| Jerk Score Vertical Acceleration during the Sit-to-Walk Transition [m] |
| Range of the Angular Velocity about Anterior-Posterior Axis during the Sit-to-Walk Transition [°/s] |
| Range of the Angular Velocity about Medio-Lateral Axis during the Sit-to-Walk Transition [°/s] |
| Range of the Angular Velocity about Vertical Axis during the Sit-to-Walk Transition [°/s] |
| Root Mean Square of the Angular Velocity about Anterior-Posterior Axis during the Sit-to-Walk Transition [°/s] |
| Root Mean Square of the Angular Velocity about Medio-Lateral Axis during the Sit-to-Walk Transition [°/s] |
| Root Mean Square of the Angular Velocity about Vertical Axis during the Sit-to-Walk Transition [°/s] |
| Normalised Jerk Score of the Angular Velocity about Anterior-Posterior Axis during the Sit-to-Walk Transition |
| Normalised Jerk Score of the Angular Velocity about Medio-Lateral Axis during the Sit-to-Walk Transition |
| Normalised Jerk Score of the Angular Velocity about Vertical Axis during the Sit-to-Walk Transition |
| Range Anterior-Posterior Acceleration during the Turn-to-Sit Transition [m/s^2] |
| Range Medio-Lateral Acceleration during the Turn-to-Sit Transition [m/s^2] |
| Range Vertical Acceleration during the Turn-to-Sit Transition [m/s^2] |
| Root Mean Square of the Anterior-Posterior Acceleration during the Turn-to-Sit Transition [m/s^2] |
| Root Mean Square of the Medio-Lateral Acceleration during the Turn-to-Sit Transition [m/s^2] |
| Root Mean Square of the Vertical Acceleration during the Turn-to-Sit Transition [m/s^2] |

| Jerk Score Anterior-Posterior Acceleration during the Turn-to-Sit Transition [m] |
| --- |
| Jerk Score Medio-Lateral Acceleration during the Turn-to-Sit Transition [m] |
| Jerk Score Vertical Acceleration during the Turn-to-Sit Transition [m] |
| Range of the Angular Velocity about Anterior-Posterior Axis during the Turn-to-Sit Transition [°/s] |
| Range of the Angular Velocity about Medio-Lateral Axis during the Turn-to-Sit Transition [°/s] |
| Range of the Angular Velocity about Vertical Axis during the Turn-to-Sit Transition [°/s] |
| Root Mean Square of the Angular Velocity about Anterior-Posterior Axis during the Turn-to-Sit Transition [°/s] |
| Root Mean Square of the Angular Velocity about Medio-Lateral Axis during the Turn-to-Sit Transition [°/s] |
| Root Mean Square of the Angular Velocity about Vertical Axis during the Turn-to-Sit Transition [°/s] |
| Normalised Jerk Score of the Angular Velocity about Anterior-Posterior Axis during the Turn-to-Sit Transition |
| Normalised Jerk Score of the Angular Velocity about Medio-Lateral Axis during the Turn-to-Sit Transition |
| Normalised Jerk Score of the Angular Velocity about Vertical Axis during the Turn-to-Sit Transition |
| Turning Angle 180° Turn [°] |
| Turning Angle of the Turn-to-Sit Phase [°] |
| Mean Angular Velocity of the 180° Turn [°/s] |
| Mean Angular Velocity of the Turn-to-Sit Phase [°/s] |
| Peak Angular Velocity of the 180° Turn [°/s] |
| Peak Angular Velocity of the Turn-to-Sit Phase [°/s] |
| Normalised Jerk Score of the 180° Turn |
| Normalised Jerk Score of the Turn-to-Sit Phase |
| Walk Duration [s] |
| Gait Speed [m/s] |
| Number of Steps in the Walk Phase (not including turns) |
| Mean Step Length [m] |
| Mean Step Duration [s] |
| Step Duration Standard Deviation [s] |
| Step Duration Coefficient of Variation [%] |
| Mean Phase Differences [°] |
| Phase Differences Standard Deviation [°] |
| Mean Phase [°] |
| Phase Standard Deviation [°] |
| Phase Coefficient of Variation [%] |
| Phase Coordination Index [%] |
| Time-Normalised Jerk Score in the Anterior-Posterior direction [m] |
| Time-Normalised Jerk Score in the Medio-Lateral direction [m] |
| Time-Normalised Jerk Score in the Vertical direction [m] |
| Normalised Jerk Score in the Anterior-Posterior direction |
| Harmonic Ratio in the Anterior-Posterior direction |
| Harmonic Ratio in the Medio-Lateral direction |

| Harmonic Ratio in the Vertical direction |
|---|
| Step Regularity in the Anterior-Posterior Direction [%] |
| Step Regularity in the Medio-Lateral Direction [%] |
| Step Regularity in the Vertical Direction [%] |
| Stride Regularity in the Anterior-Posterior Direction [%] |
| Stride Regularity in the Medio-Lateral Direction [%] |
| Stride Regularity in the Vertical Direction [%] |
| Cadence [steps/min] |
| Gait Simmetry in the Anterior-Posterior Direction |
| Gait Simmetry in the Medio-Lateral Direction |
| Gait Simmetry in the Vertical Direction |
| Range Anterior-Posterior Acceleration during the Walk Phase [m/s^2] |
| Range Medio-Lateral Acceleration during the Walk Phase [m/s^2] |
| Range Vertical Acceleration during the Walk Phase [m/s^2] |
| Root Mean Square of the Anterior-Posterior Acceleration during the Walk Phase [m/s^2] |
| Root Mean Square of the Medio-Lateral Acceleration during the Walk Phase [m/s^2] |
| Root Mean Square of the Vertical Acceleration during the Walk Phase [m/s^2] |
| Range of the Angular Velocity about Anterior-Posterior Axis during the Walk Phase [°/s] |
| Range of the Angular Velocity about Medio-Lateral Axis during the Walk Phase [°/s] |
| Range of the Angular Velocity about Vertical Axis during the Walk Phase [°/s] |
| Root Mean Square of the Angular Velocity about Anterior-Posterior Axis during the Walk Phase [°/s] |
| Root Mean Square of the Angular Velocity about Medio-Lateral Axis during the Walk Phase [°/s] |
| Root Mean Square of the Angular Velocity about Vertical Axis during the Walk Phase [°/s] |
| Number of Steps in the 180° Turn |
| Power of the Vertical Push Off in the Sit-to-Walk Transition [Nm] |
| Jerk Ratio of the Antero-Posterior Direction in the Walk Phase |
| Jerk Ratio of the Medio-Lateral Direction in the Walk Phase |

### *A.2 List of clinical features*

**Table A.2** - Full list of features obtained from clinical assessments. It is worth noting that both TUG and 10-meter walking test were repeated five times per subject and each repetition is considered independently.

| Clinical Features |
|---|
| Duration TUG (mean score over 5 repetitions) |
| Walking speed (10MWT, mean score over 5 repetitions) |
| MiniBest |
| POMAB |
| Conley |
| FES-I |
| FIM motor |
| FIM |